# Non-Interrupting Rail Track Geometry Measurement System Using UAV and LiDAR

Lihao Qiu[a], Ming Zhu[a], Yingtao Jiang[a], Hualiang (Harry) Teng[b], JeeWoong Park[b]

a. Department of Electrical and Computer Engineering, University of Nevada Las Vegas, USA 89154

b. Department of Civil and Environmental Engineering and Construction, University of Nevada Las Vegas, USA 89154

**Abstract**

The safety of train operations is largely dependent on the health of rail tracks, necessitating regular and meticulous inspection and maintenance. A significant part of such inspections involves geometric measurements of the tracks to detect any potential problems. Traditional methods for track geometry measurements, while proven to be accurate, require track closures during inspections, and consume a considerable amount of time as the inspection area grows, causing significant disruptions to regular operations. To address this challenge, this paper proposes a track geometry measurement system (TGMS) that utilizes an unmanned aerial vehicle (UAV) platform equipped with a light detection and ranging (LiDAR) sensor. Integrated with a state-of-the-art machine-learning-based computer vision algorithm, and a simultaneous localization and mapping (SLAM) algorithm, this platform can conduct track geometry inspections seamlessly over a larger area without interrupting rail operations. In particular, this semi- or fully automated measurement is found capable of measuring critical track geometry irregularities in gauge, curvature, and profile with sub-inch accuracy. Cross-level and warp are not measured due to the absence of gravity data. By eliminating operational interruptions, our system offers a more streamlined, cost-effective, and safer solution for inspecting and maintaining rail infrastructure.

## 1. Introduction

The safety of rail, as a critical component of modern national infrastructure, is paramount due to its far-reaching implications for both the economy and public well-being. Consequently, railway tracks need to be regularly inspected for potential issues. Over time, large forces at the wheel-rail contact points, along with other factors such as weather conditions, natural wear and tear, and track defects, can cause geometric irregularities in the tracks [1]. These irregularities may deviate tracks from their designed performance and compromise their integrity, directly affecting the dynamic response of railway vehicles running on those tracks [2]. This can result in disturbing noises due to increased vibrations, reduced passenger comfort due to rough rides [3], and even catastrophic derailments if left unchecked [4]. To address these challenges, a number of rail track geometry measurement technologies have emerged, many of which are now widely used in the rail industry.

The dominant methods of measuring rail track geometry use Track Geometry Measurement System (TGMS) [8], including manual measurements with specialized tools and automated measurements using various mounts and sensors. While these methods are widely utilized in the rail industry [29], they each have their limitations. Despite being accurate, manual measurements are cumbersome, labor-intensive, and inefficient, especially when assessing long stretches of track infrastructure spanning hundreds of kilometers. On the other hand, automated sensor-based TGMS provide more efficient measurements over long distances, enabling more convenient monitoring and inspection of long-range track geometry [30]. However, the TGMS requires the closure of rail sections under inspection, which interrupts normal train operations and reduces transportation efficiency.

This study aims to design a TGMS capable of scanning long stretches of track without disrupting normal rail operations. To achieve this, an unmanned aerial vehicle (UAV) is selected as the carrier for the measurement tools and electronics, and a Light Detection and Ranging (LiDAR) sensor is utilized as the major distance measurement tool. The collected data is processed via semantic segmentation for automatic rail section extraction, and using LiDAR simultaneous localization and mapping (SLAM) for comprehensive rail scene reconstruction. Polynomial regressions are used to reject outliers, and approximate the rail surface for geometry measurement, including track gauge, curvature, and profile. Cross-level and warp are not measured in this study due to absence of gravity information.

## 2. Literature review

This section introduces the track geometry parameters to measure, research in automated TGMS, and Machine Learning (ML) as well as SLAM techniques in rail related area.

### 2.1. Track geometry parameters

Common geometric irregularities of railway tracks include deviations in the distance between the rails at one location (i.e., gauge), differences in the height of the two rails at one location (i.e., cross-level), irregular vertical displacements of the rails at different locations (i.e., warp), deviations from the intended shape of the rail tops (i.e., profile or vertical curve), and curvature of the track [5]. These irregularities can be assessed by measuring key geometric parameters of the rails, allowing for maintenance and adjustments to ensure optimal track conditions.

 In measurement, gauge is the distance between the inner sides of rail heads measured at 15.875 mm (5/8 of an inch) below the top of the rail, as shown in  Fig. 1 (a). Track curvature is measured as the mid-ordinate gap between the chord (18.89 m, 62 ft) placed on the side of rail head and the rail, as depicted in Fig. 1 (b). Track profile is measured as the mid-ordinate gap between the chord (18.89 m on top of the rail) and the rail vertically, as shown in Fig. 1 (c). Cross-level is measured as vertical distance between the two rails at the same location assisted with a bubble-type level as depicted in Fig. 1 (d). Warp is measured as the cross-level between any two points within an 18.89 m span Fig. 1 (e).

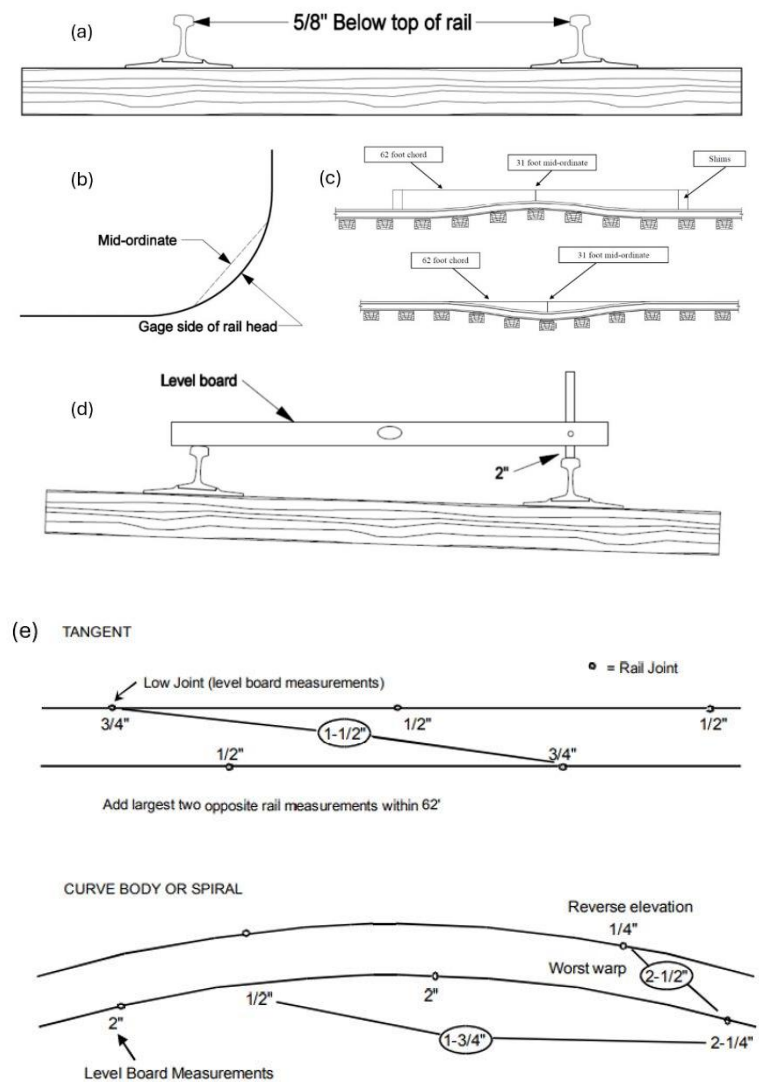

**Fig. 1.** Measurements of track geometries. (a) Gauge. (b) Curvature. (c) Profile. (d) Cross-level.(e) Warp [6].

*2.2. TGMS*

Track geometry measurement system (TGMS) use electronic devices on inspection vehicles to measure track irregularities. Common devices include inertial measurement units (IMUs), global navigation satellite systems (GNSS).

Significant research on TGMS has been conducted. For instance, Escalona et al. [33] developed a TGMS that uses kinematics from multibody dynamics to measure track irregularities accurately, showing good alignment with traditional methods. Naganuma et al. [35] introduced a trolley-mounted TGMS with gyroscopes that accurately measures seven track geometry parameters using a differential-difference method, outperforming conventional trolleys with moving parts. Another trolley-mounted ATGMS integrated with INS and geodetic surveying equipment [37], due to its modular design, offers versatile configurations for various surveying tasks.

A key advantage of these advanced TGMS is their ability to provide continuous, real-time monitoring of track conditions, allowing for early detection of defects and reducing maintenance costs. However, a significant challenge is that track inspections often require closures, disrupting rail services and creating logistical and financial issues.

*2.3. Machine learning in rail related applications*

Machine learning is a branch of artificial intelligence where algorithms learn from data to make predictions without explicit programming [34]. While ML is not widely used in TGMS, it is applied in railway data analysis, particularly railway LiDAR data. For example, Zhang et al. [16] proposed a method to reconstruct railway overhead wires from airborne LiDAR, combining deep learning for wire identification with the RANSAC algorithm for reconstruction. Lim et al. [**Error! Reference source not found.**] developed an ML-based object detection system using the SSD MobileNet model for railway safety, effectively detecting and classifying objects in images and real-time video.

*2.4. SLAM in rail related applications*

Simultaneous localization and mapping (SLAM) is used in robotics to build a map of an unknown environment while tracking the agent's location [28]. It relies on various sensors, such as cameras, LiDAR, and IMUs, and uses certain algorithms for feature extraction and alignment to integrate multiple data frames into a coherent map [32]. Though not widely adapted in TGMS, studies in the rail have explored camera and/or LiDAR based SLAM applications. For instance, Wang et al. [19] integrated data from LiDAR, camera, IMU, and GNSS on a locomotive to register point cloud data (PCD) of the rail environment with low error. Dai et al. [20] proposed a LiDAR SLAM system for mapping rail tunnels with good performance in repeated environments.

**3. Materials and methods**

*3.1. Test site*

The data collection was carried out at the Nevada State Railroad Museum located in Boulder City, Nevada, USA. An almost 210 m test section of mostly straight and slightly curved rail track was selected for this experiment, as shown in Fig. 2. This was the only track section available for experiment.

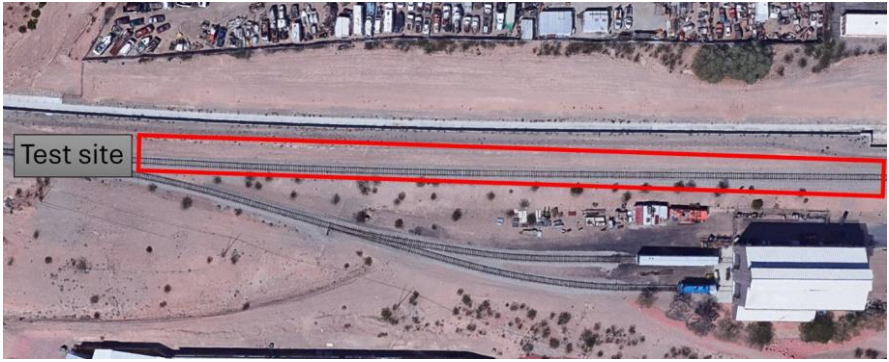

**Fig. 2.** Test site at the Nevada State Railroad Museum, Boulder City, Nevada

*3.2. Measurement equipment / Hardware*

The platform hardware includes a DJI Matrix 600 as the carrier, an Ouster OS-1 LiDAR for point cloud data generation, a 3DM-GQ7 IMU for flight status tracing, and an Nvidia Jetson Xavier Development Kit (or Jetson in short) for on-board data acquisition and processing, as shown in Fig. 3.

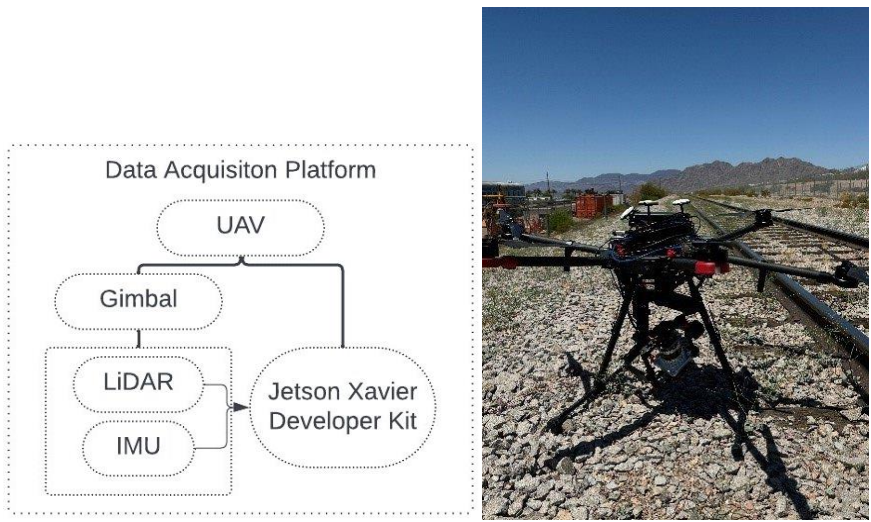

**Fig. 3.** Data acquisition platform.

The Ouster OS-1 time-of-flight (ToF) LiDAR scans with a 360° rotational field of view (FOV) horizontally and emits 128 laser channels covering 45°FOV vertically. This LiDAR outputs point cloud data (PCD) in 3D Cartesian coordinates $(x_L, y_L, z_L)$ that represent the surfaces of surround objects along with their corresponding reflection intensity measurements.

The IMU is attached on top of the LiDAR to accurately capture the instant dynamic posture of the LiDAR via high-speed accelerometers and gyroscopes. The accelerometers measure linear acceleration along three orthogonal axes $(x_I, y_I, z_I)$, while the gyroscopes measure the angular velocity around each of the three axes.

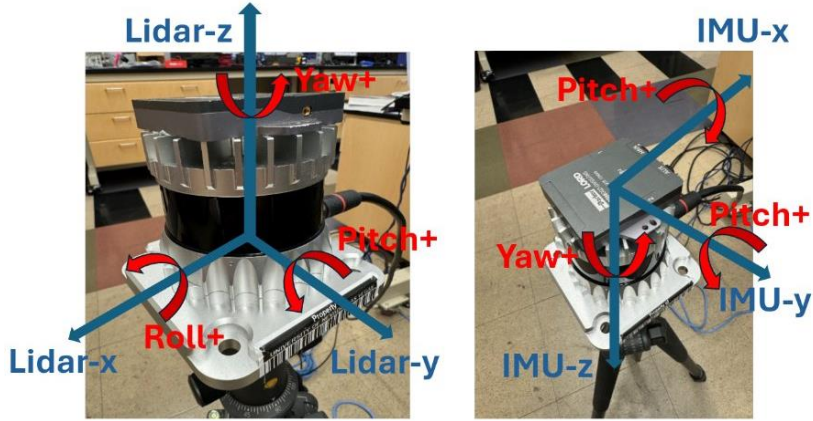

**Fig. 4.** Coordinate difference between LiDAR and IMU.

It is important to note that the 3-axes of the LiDAR are not completely aligned with the 3-axes of the IMU, as depicted in Fig. 4. Actually, the IMU coordinate system needs to be projected onto the LiDAR coordinate system using the following equations:

$$\begin{bmatrix} x_L \\ y_L \\ z_L \end{bmatrix} = \begin{bmatrix} -1 & 0 & 0 \\ 0 & 1 & 0 \\ 0 & 0 & -1 \end{bmatrix} \begin{bmatrix} x_I \\ y_I \\ z_I \end{bmatrix} \quad (1)$$

$$\begin{bmatrix} R_L \\ P_L \\ Y_L \end{bmatrix} = \begin{bmatrix} 0 & -1 & 0 \\ 1 & 0 & 0 \\ 0 & 0 & 1 \end{bmatrix} \begin{bmatrix} R_I \\ P_I \\ Y_I \end{bmatrix} \quad (2)$$

where $x$, $y$, and $z$ stands for x, y, and z-axis of each sensor, while $R$ stands for roll, $P$ stands for pitch, and $Y$ stands for yaw, respectively. The subscript $L$ and $I$ stands for LiDAR coordinate and IMU coordinate, respectively.

*3.3. Data analysis / Software*

Raw PCD frames contain both rails and irrelevant surrounding objects. To automate track geometry measurement, rail tracks must first be segmented from background objects like buildings, trees, even grounds, and so on. The segmented PCDs with small vertical FOV are then registered for large-scale curvature and profile measurement. Outliers are removed to eliminate misclassified points from segmentation. Finally, various track geometry parameters are calculated. The overall flow of the data analysis process is shown in Fig. **5**.

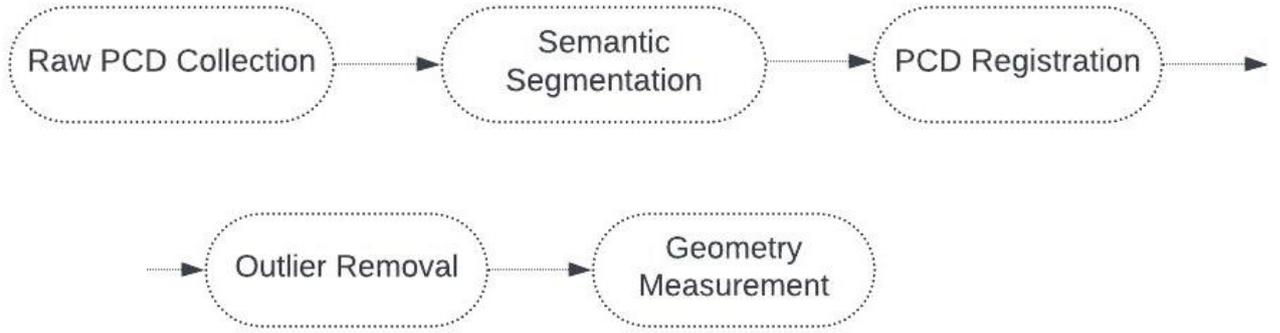

**Fig. 5.** Flowchart for data processing

*3.3.1.    Semantic segmentation*

To enable automatic segmentation of rail point clouds across different scenarios, an ML approach is employed. Firstly, the collected raw PCD frames were labeled using Supervisely [21] (Fig. 6 (a)), where a cuboid was positioned to encapsulate points representing rails. Unlabeled points are classified as background. A total of 358 PCD frames were labeled as ground truth. As most data contained two tracks, cropping was applied to retain rail points and remove background, as shown in Fig. 6 (b). Each PCD frame was reduced to 50,000 points nearest to the LiDAR, ensuring all frames contained an equal number of points for consistent ML training, while increasing the rail points-background points ratio.

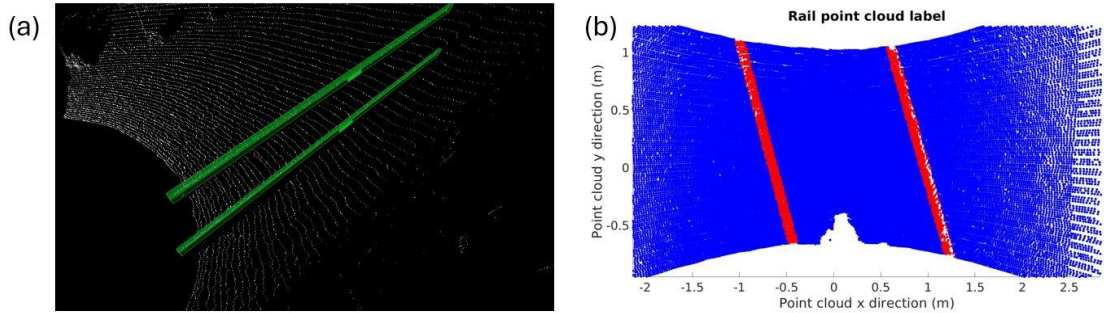

Fig. 6. (a)Using cuboid to annotate rail points. (b) Annotated and cropped rail point cloud frame. Red points are labeled rail points.

To increase the diversity of training samples, data augmentation techniques such as rotation, translation and noises were applied to the original labeled frames [22]. Eventually, 358 labeled PCD frames were augmented to a total of 1,322 frames for the entire rail PCD dataset. The dataset was further split with a 0.7:0.2:0.1 ratio for training, validation, and testing purposes, respectively.

A ML-based semantic segmentation neural network [23] is optimized for rail point segmentation. The network uses an encoder-decoder structure with skip connections. During the encoding process, random sampling downsamples the points, while local spatial encoding (LocSE) modules and attentive pooling (AP) modules aggregate point features to retain information from unselected points. The LocSE module selects a point $p_i$ with feature $f_i$ and calculate its $K$ neighboring points $[p_i^1, p_i^2, \ldots, p_i^K]$ in terms of their relative point position information $r_i^k$:

$$r_i^k = MLP(p_i \oplus p_i^k \oplus (p_i - p_i^k) \oplus ||p_i - p_i^k||) \tag{3}$$

where $k = [1, 2, \ldots, K]$, $MLP$ stands for multi-layer perceptron, $\oplus$ stands for concatenation, and $||\cdot||$ calculates the Euclidean distance. The information $r_i^k$ are then concatenated to the feature of point $p_i$ (i.e., $\hat{f}_i^k = f_i \oplus r_i^k$), forming features $\hat{F}_i = [\hat{f}_i^1, \hat{f}_i^2, \ldots, \hat{f}_i^K]$. The AP module aggregates the set of neighboring point features $\hat{F}_i$. The attention score $s_i^k$ is computed through a shared $MLP$ layer followed by $softmax$ function:

$$s_i^k = softmax(MLP(\hat{f}_i^k, W)) \tag{4}$$

where $W$ is the learnable weights for the shared $MLP$ layers. The weighted features are then calculated as:

$$\tilde{f}_i = \sum_{k=1}^{K}(\hat{f}_i^k \cdot s_i^k) \tag{5}$$

In the decoder stage, up-sampling modules and multi-layer perceptron (MLP) are used to predict the label for each point in the original PCD frame. Cross entropy with logits loss function:

$$loss_i = -\sum_{j=0}^{1} y_{i,j} \cdot ln p_{i,j} \tag{6}$$

is employed to compute the loss between predicted labels of each point cloud and its ground truth, where $y_{ij}$ stands for the ground truth for the $j$ th point in the $i$ th frame, and $p_{ij}$ stands for the prediction for the $j$ th point in the $i$ th frame. The final result is achieved by minimizing all losses.

The annotated dataset faces a significant issue of data imbalance, with rail points comprising only 1-2% of the total samples, as shown in Fig. 6 (b). To mitigate this imbalance, modifications were made to both the downsampling rates and feature dimensions. Specifically, the downsampling ratio in the second and third random sampling stages was doubled compared to the original network. Additionally, the feature dimensions after each local feature aggregation module were halved. The overall network structure used is illustrated in Fig. 7.

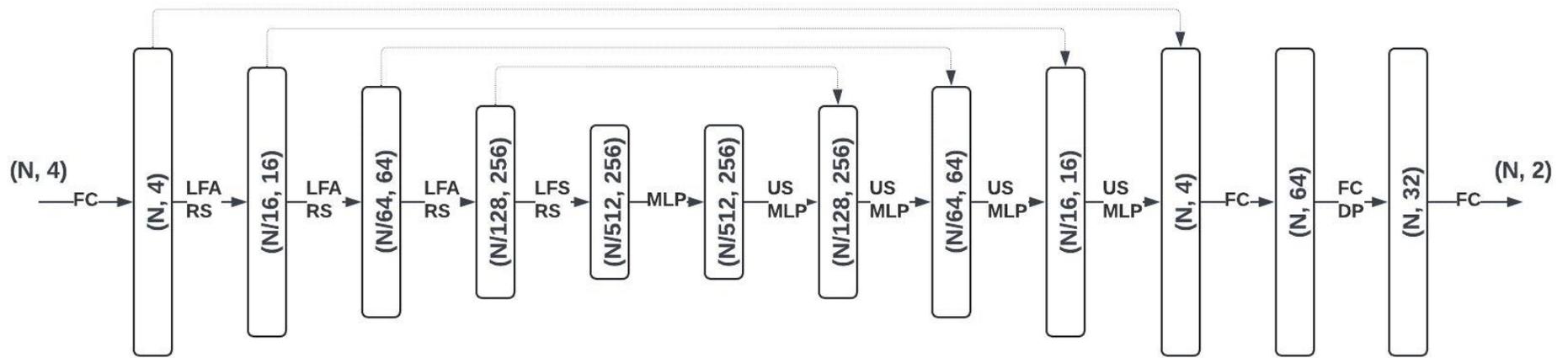

**Fig. 7.** Network structure. (N, F) stands for N points and F dimension of features. FC: fully connected layer. LFA: local feature aggregation. LFA consists of multiple LocSE and AP modules. RS: random sampling. MLP: multi-layer perceptron. US: up-sampling. DP: dropout layer.

*3.3.2.    Point cloud registration*

For track curvature and profile measurements over lengths of at least 18.89 meters, the FOV of a single PCD frame does not suffice. To address this, multiple PCD frames are registered into a unified coordinate system to create a larger scene using a SLAM algorithm [24] relying on IMU kinematic readings and point cloud features.

*3.3.2.1.    IMU Kinematics*

Raw IMU readings are presented as follows:

$$\hat{w}_t = w_t + b_t^w + n_t^w \tag{7}$$
$$\hat{a}_t = R_t^{BW}(a_t - g) + b_t^a + n_t^a \tag{8}$$

where $\hat{w}_t$ and $\hat{a}_t$ represent the noisy angular velocity and acceleration readings from the IMU in world frame at time $t$. These readings are interfered by a slowly varying bias $b_t$ and white noise $n_t$. $R_t$ stands for the rotation matrix from world coordinate $W$ system to body coordinate system $B$, and $g$ stands for the constant gravity vector in world coordinate system.

The velocity, position, and rotation of the LiDAR at time $t + \Delta t$ can be computed as follows:

$$v_{t+\Delta t} = v_t + g\Delta t + R_t(\hat{a}_t - b_t^a + n_t^a)\Delta t \tag{9}$$
$$p_{t+\Delta t} = p_t + v_t\Delta t + \frac{1}{2}g\Delta t^2 + \frac{1}{2}R_t(\hat{a}_t - b_t^a + n_t^a)\Delta t^2 \tag{10}$$
$$R_{t+\Delta t} = R_t^{WB}\exp\left((\hat{w}_t - b_t^w - n_t^w)\Delta t\right) \tag{11}$$

The preintegrated IMU method, as described in [25], is then applied to obtain the incremental body motion (i.e., velocity, position, rotation) between two timestamps:

$$\Delta v_{i,j} = R_i^T(v_j - v_i - g\Delta t_{i,j}) \tag{12}$$
$$\Delta p_{i,j} = R_i^T(p_j - p_i - v_i\Delta t_{i,j} - \frac{1}{2}g\Delta t_{i,j}^2) \tag{13}$$
$$\Delta R_{i,j} = R_i^T R_j \tag{14}$$

These values are jointly updated with point cloud feature matching.

*3.3.2.2.    Point Cloud Features*

Point cloud features are selected by calculating the roughness of each point $p_{i,k}$ ($k$ th point in the $i$ th frame):

$$c_{i,k} = \frac{1}{|S|\cdot\|p_{i,k}\|} \left\| \sum_{j\in S, j\neq k}(p_{i,k} - p_{i,j}) \right\| \tag{15}$$

where $S$ stands for the set of points surrounding $p_{i,k}$ on the same horizontal channel, with an equal number of points on each side of $p_{i,k}$. Points with high roughness are considered as edge features, and points with low roughness are treated as planar features. A voxel map containing features from previous n frames is used for feature matching. A new scan $F_{i+1}$ with edge feature $F_{i+1}^e$ and planar feature $F_{i+1}^p$ is matched to the voxel map through planar to planar, edge to edge matching:

$$d_{e_k} = \frac{|(p_{i+1,k}^e - p_{i,u}^e) \times (p_{i+1,k}^e - p_{i,v}^e)|}{|p_{i,u}^e - p_{i,v}^e|} \tag{16}$$

$$d_{p_k} = \frac{\left|\left(p_{i+1,k}^p - p_{i,u}^p\right) \left(p_{i,u}^p - p_{i,v}^p\right) \times \left(p_{i,u}^p - p_{i,w}^p\right)\right|}{|(p_{i,u}^p - p_{i,v}^p) \times (p_{i,u}^p - p_{i,w}^p)|} \tag{17}$$

where $d_{e_k}$ represents the distance between the edge point $p_{i+1,k}^e$ to a line formed by the corresponding edge points $p_{i,u}^e$ and $p_{i,v}^e$, $d_{p_k}$ represents the distance between the planar point $p_{i+1,k}^p$ and the plane formed by corresponding planar points $p_{i,u}^p$, $p_{i,v}^p$, and $p_{i,w}^p$. Gauss-Newton method is used to solve for the optimal transformation by minimizing the combined distance among all edge and planar features:

$$\min_{T_{i+1}} \{\sum_{p_{i+1,k}^e \in \acute{F}_{i+1}^e} d_{e_k} + \sum_{p_{i+1,k}^p \in \acute{F}_{i+1}^p} d_{p_k}\} \tag{18}$$

where $\acute{F}_{i+1}$ represents features transformed with preintegrated relative pose values.

### 3.3.3. Outlier rejection

After PCD segmentation and SLAM are applied, mislabeled rail point should be removed from the registered rail landscape. To do so, all rail points are rotated using principal component analysis (PCA) to align the rail's longitudinal direction with the x-axis, track width with the y-axis, and height with the z-axis. The rails are then divided into 5-meter segments for a more accurate regression fitting. Each segment is re-aligned in the same way using PCA and fitted with a quadratic regression

$$y = ax^2 + bx + c \tag{19}$$

where a, b, and c are coefficients, is used for fitting rail points in the x-y plane, and another quadratic regression for the x-z plane to approximate each track respectively. The distance of each point to these two regressions in each plane is then calculated. Points with distance greater than a determined threshold are removed.

### 3.3.4. Track Geometry measurement

Manual geometry measurement requires identifying specific points or surfaces on the rail. To achieve this, polynomial regressions are employed to determine the rail surfaces in the registered PCD map for measuring gauge, curvature, and profile. Additionally, a sliding window method is used to select critical measurement points and to obtain more continuous measurement.

### 3.3.4.1. Track gauge measurement

To simulate track gauge measurement, the rail head is first identified, and the inner rail surface is located 15.875 mm (i.e., 5/8 of an inch) below it. The top 10% of points in each segment are used to approximate the rail head, and the innermost 10% of points that are 15.75 - 16 mm below the top are selected as inner surface points. Due to the sparse PCD data, direct measurement is difficult, so linear regressions are fitted to interpolate these points. A 5-meter long sliding window is used to select the critical points, where $(x_{1,1}, y_{1,1}), (x_{1,2}, y_{1,2})$ are two points on one linear regression line that are 5 meters apart, and $(x_{2,1}, y_{2,1}), (x_{2,2}, y_{2,2})$ are the ones 5 meters apart on the other linear regression line (Fig. 8). The sliding gauge value is then calculated as the distance between two midpoints of both linear regression line segments (i.e., $(x_{1,m}, y_{1,m}) = ((x_{1,1} + x_{1,2})/2, (y_{1,1}, y_{1,2})/2)$ and so forth) [26]:

$$d_g = \sqrt{(x_{1,m} - x_{2,m})^2 + (y_{1,m} - y_{2,m})^2} \tag{20}$$

and the window shifts 0.5 m at a time to minimize non-linearity and ensure continuity.

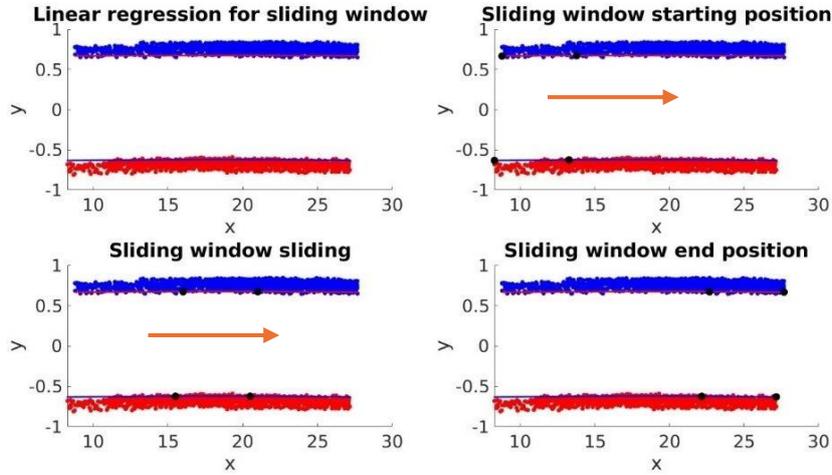

**Fig. 8.** Sliding window demonstration. TL: linear regression line for gauge. TR: five-meter sliding window starting point. BL: sliding window in process. BR: sliding window stop position.

*3.3.4.2. Track curvature measurement*

To accurately measure track curvature, points along the gauge side of the rail head should be used to approximate the manual measurement method (Fig. 1 (b)). Nevertheless, due to the sparse nature of PCD, this approach can lead to inaccurate regression fitting. Instead, all rail head points within an 18.89 m-long window are used for regression fitting in the x-y plane.

The top 10% of rail points in each window are selected and fitted with a quadratic regression, representing the gauge side. Linear regression is applied to points at both ends to approximate the 62-foot chord. Curvature is then calculated as the lateral distance between the quadratic and linear regressions at the window's midpoint. The window slides 3 m at a time for continuous curvature measurement.

*3.3.4.3. Track Profile measurement*

For track profile measurement, the top 10% of rail head points in each window are selected. In the x-z plane, these points are fitted with a quadratic curve, while straight lines are fitted at both ends. The profile is measured by calculating the vertical distance between the quadratic curve and the linear fits at the window's midpoint. This process is repeated every 3 meters for continuous profile measurement.

## 4. Results

*4.1. Semantic segmentation result*

The precision of the track geometry calculations depends heavily on the accuracy of the semantic segmentation process. During the training process, the batch size is set to eight, and the learning rate is set to 0.01. Given the significant imbalance between the two classes of points in rail PCD, the Intersection over Union (IoU) metric is employed to assess the model's performance, rather than traditional metrics such as accuracy or F1-score. After the training and validation process, the model demonstrated a highest IoU of 82.4%.

*4.2. Point cloud registration result*

The process of point cloud registration involves aligning and merging successive LiDAR frames to create a comprehensive, detailed composite representation of the test site. Fig. 9 presents the registered point cloud data of the test site. This figure showcases the effectiveness of the registration process, with the rails prominently highlighted in red. This coloration is based on the predicted labels, clearly distinguishing the rails from other elements within the scene.

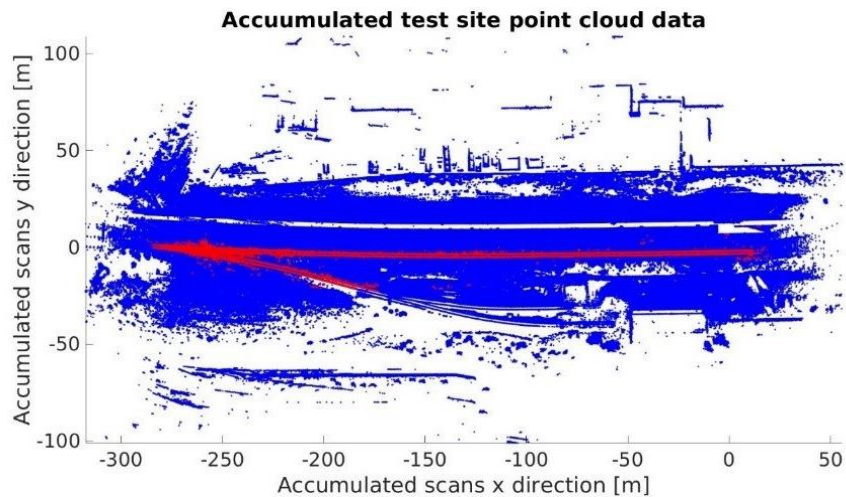

**Fig. 9.** Registration result

*4.3. Outlier rejection result*

Outliers in the registered map can significantly affect measurement accuracy. To address this, outliers are removed using the distance-based method aforementioned. Specifically, thresholds of 7 cm in the x-y plane and 8 cm in the x-z plane are applied based on visual inspection post-outlier removal. In this dataset, 10,511 outliers were identified and removed from a total of 50,503 rail points.

*4.4. Geometry measurement result*

All measurement results are compared to field measurements taken with specialized rail tools, as shown in Fig. 10. The track gauge was measured every 5 meters, while track curvature and profile measurements were taken 18.89 meters (62 feet) apart, totaling 11 measurements. All field measurements began at a landmark position indicated in Fig. 10. The platform's measurement results are compared to the field results when each sliding window location aligns with the corresponding on-field measurement location.

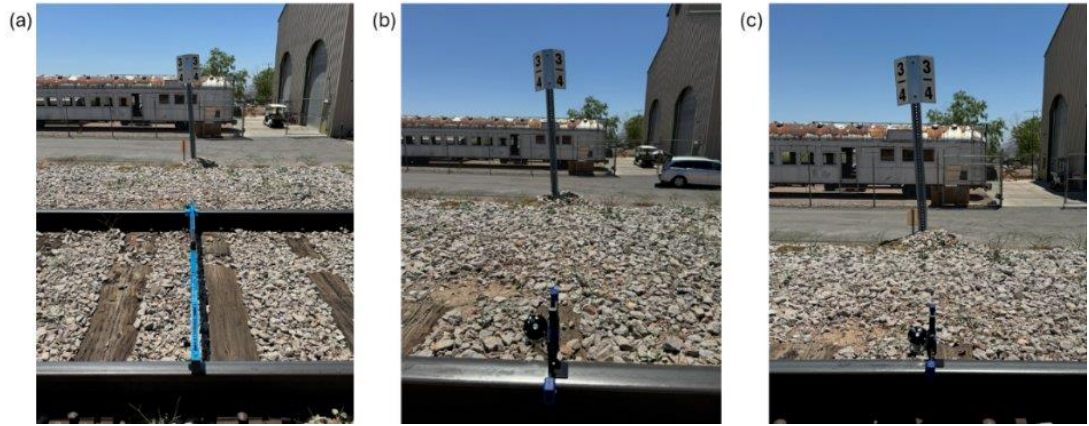

**Fig. 10.** Rail Measurement. (a) Measuring gauge from the landmark using gauge measuring tool. (b) Measuring curvature from the landmark using curvature measuring tool. (c) Measuring profile from the landmark using profile measuring tool.

*4.4.1. Track Gauge*

Fig. 11 illustrates the gauge measurement process in the PCD that simulates this measurement method. The inner surface is identified by locating it approximately 15.75-16 mm below the rail head. Fig. 14 (a) presents a histogram of deviations between these two sets of gauge measurements. Notably, around 78.57% of these deviations are within 2cm.

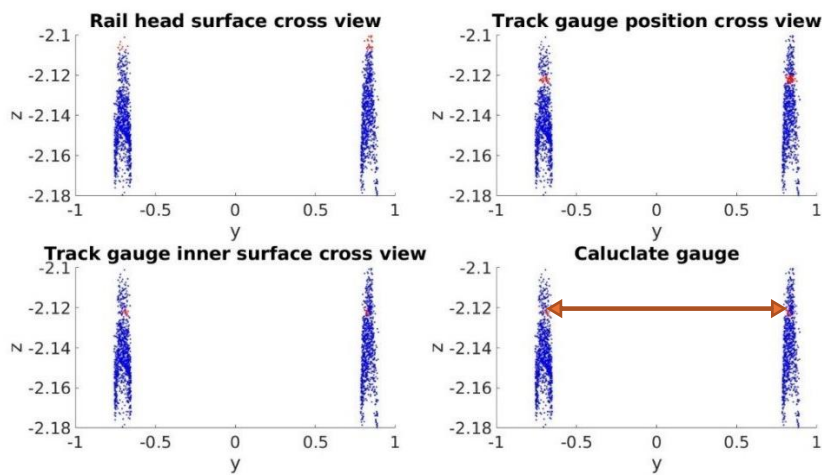

**Fig. 11.** Gauge calculation process. (TL) Selecting rail head. (TR) Finding 15.75 – 16 mm below rail head. (BL) Finding inner surface. (BR) Calculating gauge.

*4.4.2.  Track Curvature*

Fig. 12 illustrates the curvature measurement process that simulates this method by fitting a linear regression as the chord and a quadratic regression at the gauge side of the reference rail in the x-y plane. Fig. 14 (b) shows the histogram depicting the deviations between the measured curvature and the calculated curvature values. Overall, 77.27% of the deviations fall within 1 cm.

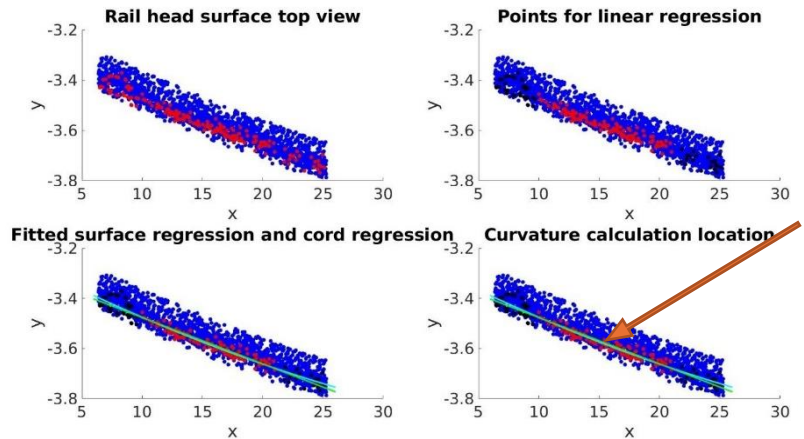

**Fig. 12.** Track curvature calculation process. (TL) Finding rail head points. (TR) Finding the points on both ends for chord simulation. (BL) Fitting the surface regression and the cord regression. (BR) Calculating the gap at the midpoint.

*4.4.3.  Track Profile*

Fig. 13 demonstrates the profile measurement in the rail points through fitting a linear regression as the string line and a quadratic regression as the top of the reference rail in the $x - z$ plane. Fig. 14 (c) depicts a histogram of the deviations between the measured profile values and the calculated profile values, where 77.27% of the deviations fall within 1 cm.

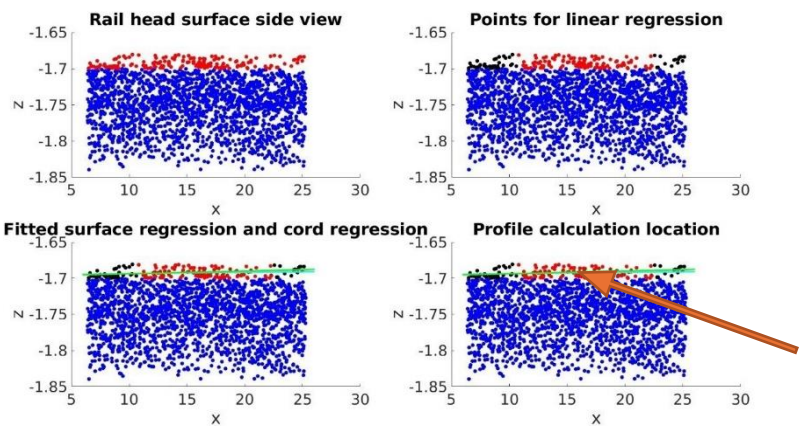

**Fig. 13.** Profile calculation. (TL) Finding rail head. (TR) Finding chord location. (BL) Fitting rail head and chord. (BR) Calculating profile gap

**Table 1**
Part of ground truth and measurement result. GT: ground truth. Mt.: measurement. L: Left rail. R: Right rail. Left rail is the one to the south. All units are shown in cm.

| Case | Gauge GT | Gauge Mt. | Curvature GT (L/R) | Curvature Mt. (L/R) | Profile GT (L/R) | Profile Mt. (L/R) |
|---|---|---|---|---|---|---|
| 1 | 143.5 | 144.5 | 0.15/0.47 | 0.37/0.12 | 0.80/0.00 | 1.15/0.00 |
| 2 | 143.8 | 143.5 | 2.06/2.38 | 2.04/2.00 | 0.60/0.90 | 0.00/0.85 |
| 3 | 143.5 | 143.5 | 1.91/2.06 | 1.94/1.22 | 1.40/0.90 | 0.13/0.14 |
| 4 | 144.1 | 145.2 | 0.64/0.64 | 0.41/0.07 | 1.90/1.50 | 0.00/0.29 |
| 5 | 144.5 | 144.2 | 0.16/0.16 | 0.70/0.63 | 0.20/0.60 | 0.85/0.51 |

Table 3 presents a sample of the ground truth and measured results. All measurements are in unit centimeter (cm).

Table **2** presents the mean, standard deviation (Std), root mean square error (RMSE), and average relative error percentage (AREP) for all three calculations. While the RMSE values are similar, the gauge measurement shows a significantly lower AREP. The measured gauge values are around 1.435 m, resulting in a relative error of approximately 0.77%. In comparison, the relative error percentages for curvature and profile are notably higher.

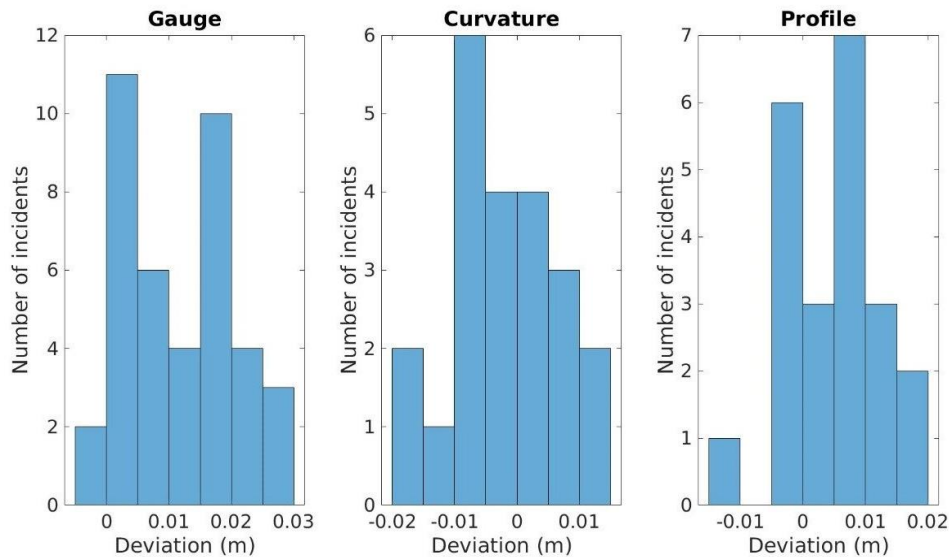

**Fig. 14.** Histogram of: (a) Gauge difference. (b) Curvature difference. (c) Profile difference.

**Table 2**
Platform calculation result summary.

| Parameter | Mean (cm) | Std (cm) | RMSE (cm) | AREP |
|---|---|---|---|---|
| Gauge | 144.88 | 1.68 | 1 | 0.77% |
| Curvature | 0.97 | 0.84 | 0.84 | 143.77% |
| Profile | 0.124 | 0.35 | 0.87 | 111.20% |

## 5. Discussion

The results of this study demonstrate the proposed system's potential for measuring railroad track geometries. Specifically, the track gauge measurement showed high accuracy, with an RMSE of 1 cm and an AREP of 0.77%. However, larger deviations were observed in measuring track curvature and profile, with AREP values exceeding 100%. This may be due to the test tracks being mostly straight, resulting in minimal curvature or profile variation. Testing on curved sections could yield different results.

Currently, the system relies only on point cloud data and lacks information about gravity direction at each point, preventing accurate measurement of cross-level and warp values. Future improvements should focus on integrating sensors or algorithms to capture these parameters, enhancing the precision of measurements and advancing the system's capabilities in track geometry measurement.

## 6. Conclusion

This paper presented an innovative rail track geometry measurement platform capable of conducting inspections alongside normal rail operations. Utilizing a UAV platform equipped with a LiDAR sensor and an on-board data acquisition system, the proposed approach integrates a state-of-the-art machine-learning-based computer vision algorithm for rail point segmentation and a LiDAR SLAM algorithm for expanding the point cloud field of view. Through the application of regression techniques for outlier removal and precise geometry calculations, the platform has demonstrated high accuracy in measuring critical track geometry parameters, including gauge, curvature, and profile, with sub-inch precision. Compared to traditional field measurements using specialized tools, our system significantly reduces operational interruptions, offering a more streamlined, cost-effective, and safer solution for inspecting and maintaining rail infrastructure. Future work will focus on enhancing measurement accuracy further and incorporating additional assessments such as cross-level and warp evaluations to broaden the platform's capabilities.

**Acknowledgement**

The authors would like to thank the Nevada State Railroad Museum for the test site they provided and their collaboration. This work was supported by U.S. Department of Transportation University Transportation Program with contract number 69A3551747132.

**Declaration of conflicting interest**

The author(s) declared no potential conflicts of interest with respect to the research, authorship, and/or publication of this article.